\documentclass[acmconf, authorversion]{acmart}

\usepackage{url}
\usepackage{hyperref}

\usepackage{graphicx}
\usepackage{multirow}
\usepackage{xcolor}
\usepackage{xspace}
\usepackage{enumitem}
\usepackage{tabularx}
\usepackage{booktabs}
\usepackage{diagbox}

\DeclareGraphicsExtensions{.pdf,.png,.jpg,.jpeg}
\graphicspath{
    {figures/}
    {images/}
    {./}
}

\usepackage{array}
\newcolumntype{L}[1]{>{\raggedright\let\newline\\\arraybackslash\hspace{0pt}}m{#1}}
\newcolumntype{C}[1]{>{\centering\let\newline\\\arraybackslash\hspace{0pt}}m{#1}}
\newcolumntype{R}[1]{>{\raggedleft\let\newline\\\arraybackslash\hspace{0pt}}m{#1}}
\newcolumntype{P}[1]{>{\raggedright}p{#1}}

\AtBeginDocument{%
  \providecommand\BibTeX{{%
    \normalfont B\kern-0.5em{\scshape i\kern-0.25em b}\kern-0.8em\TeX}}}

\copyrightyear{2021}
\acmYear{2021}
\acmDOI{}

\acmConference[KDD Workshop Fragile Earth '21]{Fragile Earth '21: KDD 2021 Workshop}{August 14--18, 2021}{Virtual}
\acmBooktitle{Fragile Earth '21: KDD 2021 Workshop}

\begin{document}

%%
%% The "title" command has an optional parameter,
%% allowing the author to define a "short title" to be used in page headers.
\title[Model Generalization in Land Cover Mapping]{Model Generalization in Deep Learning Applications for Land Cover Mapping}

%%
%% The "author" command and its associated commands are used to define
%% the authors and their affiliations.
%% Of note is the shared affiliation of the first two authors, and the
%% "authornote" and "authornotemark" commands
%% used to denote shared contribution to the research.
\author{Lucas Hu}
\affiliation{%
  \institution{University of Southern California}
  \country{USA}
}
\email{lucashu@usc.edu}

\author{Caleb Robinson}
\affiliation{%
  \institution{Microsoft, AI for Good}
  \country{USA}
}

\author{Bistra Dilkina}
\email{dilkina@usc.edu}
\affiliation{%
 \institution{University of Southern California, Center for AI in Society}
 \country{USA}
}

%%
%% By default, the full list of authors will be used in the page
%% headers. Often, this list is too long, and will overlap
%% other information printed in the page headers. This command allows
%% the author to define a more concise list
%% of authors' names for this purpose.
\renewcommand{\shortauthors}{Hu et al.}

%%
%% The abstract is a short summary of the work to be presented in the
%% article.
\begin{abstract}
  Recent work has shown that deep learning models can be used to classify land-use data from geospatial satellite imagery. We show that when these deep learning models are trained on data from specific continents/seasons, there is a high degree of variability in model performance on out-of-sample continents/seasons. This suggests that just because a model accurately predicts land-use classes in one continent or season does not mean that the model will accurately predict land-use classes in a different continent or season. We then use clustering techniques on satellite imagery from different continents to visualize the differences in landscapes that make geospatial generalization particularly difficult, and summarize our takeaways for future satellite imagery-related applications.
\end{abstract}

%%
%% The code below is generated by the tool at http://dl.acm.org/ccs.cfm.
%% Please copy and paste the code instead of the example below.
%%


%%
%% Keywords. The author(s) should pick words that accurately describe
%% the work being presented. Separate the keywords with commas.
%\keywords{deep learning, generalization, land cover, satellite imagery}

%% A "teaser" image appears between the author and affiliation
%% information and the body of the document, and typically spans the
%% page.
\begin{teaserfigure}
    \centering
    \includegraphics[width=0.8\textwidth]{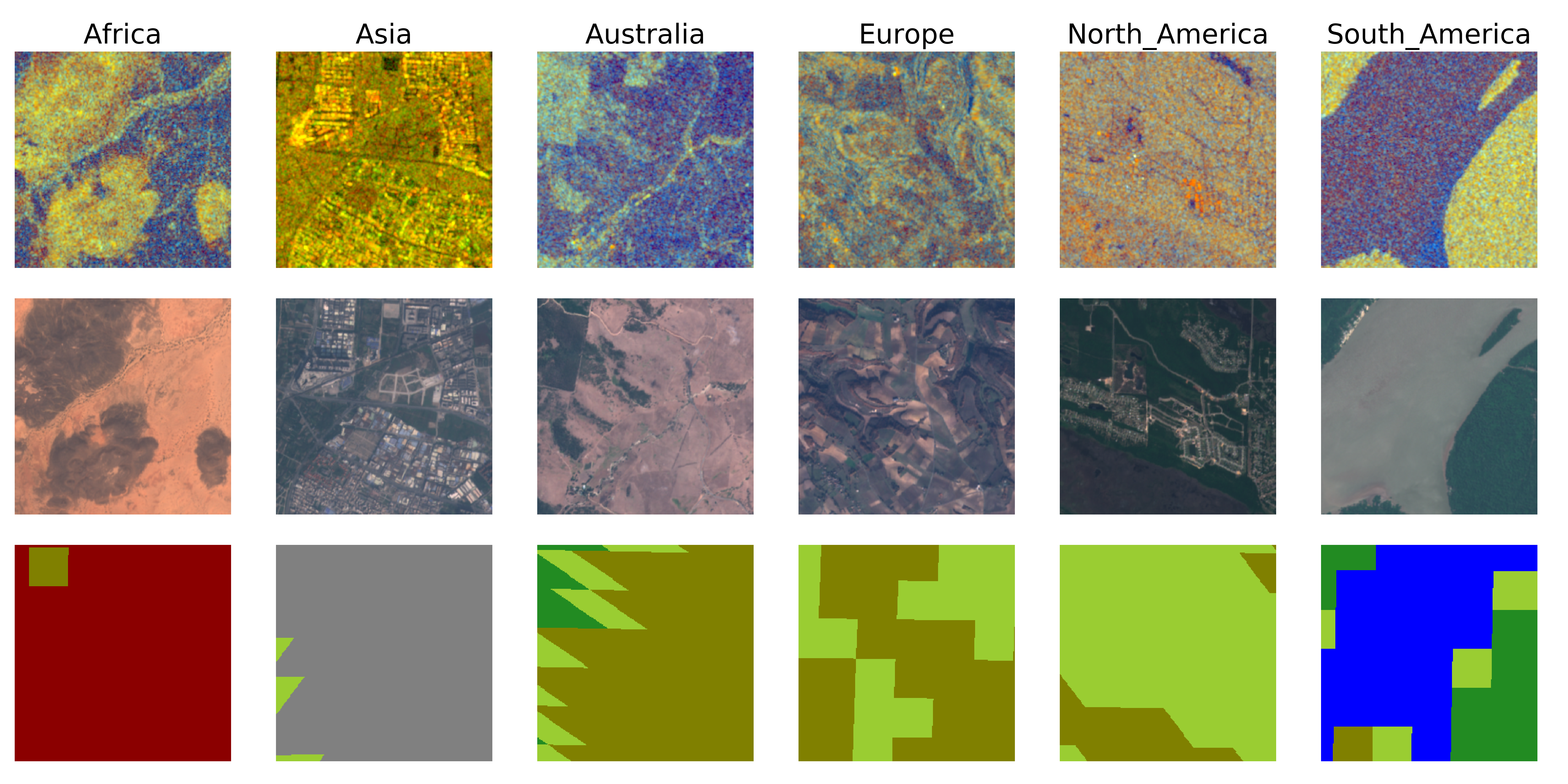}
    \captionsetup{justification=centering}
    \vspace{-2mm}
    \caption{Sample from each continent. Top: False color Sentinel-1 SAR. Middle: Sentinel-2 RGB. Bottom: Consolidated LCCS land-use.}
    %    \caption{Sample imagery from each continent. Top: False color Sentinel-1 SAR (Red channel: VV, Green channel: VH, Blue channel: VV/VH). Middle: Sentinel-2 RGB. Bottom: Consolidated LCCS land-use (LU).}
    \label{fig:sample-images}
\end{teaserfigure}

%%
%% This command processes the author and affiliation and title
%% information and builds the first part of the formatted document.
\maketitle

\section{Introduction}
The UN has estimated that, ``approximately 20\% of the SDG indicators can be interpreted and measured either through direct use of geospatial data itself or through integration with statistical data''~\cite{arnoldglobal}.  For example, land use/land cover (LULC) data -- a type of geospatial data --  is useful in many downstream sustainability applications. Conservation biologists can use land cover data to target the creation of riparian forest buffers -- areas that protect streams from collecting pollutants from adjacent areas -- an important conservation task that can improve the overall ``health'' of a watershed. As a specific example, the Chesapeake Bay Conservancy uses, ``flow path data and high-resolution land cover data to identify opportunity areas for planting riparian forest buffers within a specified distance of the flow paths. Once these restoration opportunity areas (ROAs) are identified, they can be characterized by the land cover composition and modeled sediment and nutrient loading of the upstream land area that drains through the ROA''~\cite{chesapeake}.

LULC data can be categorized as ``geospatial'' data, however, more specifically, it is a product of satellite or aerial imagery -- it must be generated through a combination of modeling and human labeling efforts from ``raw'' geospatial data. In existing GIS workflows, the process of generating land cover data is traditionally a semi-automated one whereby models are used to make initial land cover predictions from imagery, then, in a laborious process, human experts are used to correct the output of the model. Improvements in the modeling step (imagery to land cover) translate directly to savings in human labor, and indirectly into better land cover products. With the success of deep learning -- specifically convolutional neural networks -- it is natural to cast modeling problems with geospatial data -- like ``generating land cover maps from satellite imagery'' -- as vision problems, then apply state of the art methods on them~\cite{8127684,cnnremote,8024178}. This strategy poses several problems from a machine learning perspective, though:

First, most existing state of the art computer vision methods have been developed for natural images (e.g. pictures taken with a camera) and may be suboptimal when applied to geospatial images. Existing methods have specific inductive biases that are transferred to the geospatial modeling problem as entire network architectures are copied. In at least 2 of the top papers (by performance) published in the CVPR 2018 Deep Globe workshop~\cite{8575485}, the authors relied on ResNet architectures without modifications. As an example of how this may cause problems: the number of channels in natural images is 3 (red, green, and blue), while multispectral satellite imagery can have $>10$ channels. Filters in the first convolutional layer of a CNN will learn ``low-level'' features of the input, such as edges and colors, but common network architectures -- again, tuned for natural images -- contain relatively few filters. For example, the ResNet family of networks uses 64 filters in the first convolutional layer, while the Inception family uses 32 filters. This number of filters appears to be sufficient for RGB inputs, but it may well be a performance bottleneck for multispectral inputs, where more filters may be needed to fully capture features outside of the visible spectrum.

Second, ML model generalization is usually measured on a held-out dataset that is assumed to be drawn from the same distribution as the training data set. With geospatial data, this assumption can be violated in many ways. Imagery from a given satellite can vary based on: location of the image, time of day, time of year, atmospheric conditions (e.g. clouds), and geolocation errors (the error in the geolocation of a specific pixel may be many meters). Similarly, imagery of the same location/date but from different satellites can vary based on: spatial resolution of the satellites (e.g. 10 meters/px versus 1m/px), spectral resolution of the satellites (e.g. which wavelengths of light are captured in each band), and radiometric resolution of the satellites (e.g. a channel may be recorded in $[0,2^8)$ or $[0,2^{16})$ based on how sensitive the sensors are to fluctuations in light intensity). Importantly, these modes of variation are all documented in most geospatial data sources, and as such, their effects on model generalization can be studied in a controlled setting. A model trained for the previously mentioned problem (imagery → land cover) may only generalize to areas immediately surrounding the location on which it was trained, or may only generalize to imagery captured during the same season that the training imagery was captured in. Explaining how a geospatial model generalizes (and thus when additional techniques may be needed to expand the capabilities of the model) is a valuable exercise to show the ML community.

Therefore, we seek to answer three main research questions:
\begin{enumerate}
    \item How well do models trained on one season generalize to other seasons?
    \item How well do models trained on one continent generalize to other continents?
    \item What features or indicators might we use to predict when model generalization will be poor?
\end{enumerate}

\section{Dataset} \label{dataset}
We test our three research questions with the  recently released SEN12MS dataset~\cite{schmitt2019sen12ms}, which is made up of satellite imagery sampled from various scenes from across the globe. Each scene contains data from one particular region on Earth, from a particular season. There are 252 total scenes in the dataset. 
Each scene is composed of hundreds of ``patches'' of satellite imagery (each scene can contain a variable number of patches). A single ``patch'' represents a 2.56km x 2.56km plot of land in the real world. Each patch is represented in the dataset as a 256x256 pixel image, meaning that each pixel corresponds to a 10m x 10m plot of land in the real world. (For this reason, we say that the dataset has a 10m spatial resolution.)
In total, the SEN12MS dataset has 180,662 such patches across its 252 scenes. The patches collectively occupy about 512GB of disk space. A per-continent and per-season breakdown of these scene/patch counts is shown in Table ~\ref{tbl:scene-patch-counts}.

\begin{table}[!htb]
\centering
 \begin{tabular}{|p{2cm}|p{1.5cm}|p{1.5cm}|p{1.5cm}|p{1.5cm}|p{1.5cm}|}
 \hline
 \backslashbox[2.3cm]{Continent}{Season} & Spring & Summer & Fall & Winter & \textbf{Total} \\
 \hline
 Africa & 13 (9,393) & 4 (2,356) & 16 (11,776) & 11 (7,260) & \textbf{44 (30,785)} \\
 \hline
 Asia & 11 (8,438) & 16 (10,815) & 20 (14,565) & 12 (8,623) & \textbf{59 (42,441)} \\
 \hline
 Australia & 2 (1,547) & 6 (4,441) & 3 (2,279) & 2 (1,331) & \textbf{13 (9,598)} \\
 \hline
 Europe & 14 (10,284) & 19 (13,788) & 19 (13,994) & 8 (5,809) & \textbf{60 (43,875)} \\
 \hline
 N. America & 12 (8,334) & 14 (10,392) & 21 (15,484) & 10 (6,598) & \textbf{57 (40,808)} \\
 \hline
 S. America & 4 (2,887) & 6 (3,961) & 6 (4,103) & 3 (2,204) & \textbf{19 (13,155)} \\
 \hline
 \textbf{Total} & \textbf{56 (40,883)} & \textbf{65 (45,753)} & \textbf{85 (62,201)} & \textbf{46 (31,825)} & - \\
 \hline
\end{tabular}
\caption{SEN12MS scene (\& patch) counts.}
\label{tbl:scene-patch-counts}
\vspace{-5mm}
\end{table}

Each pixel in each patch contains data from three distinct sources:
(1) Sentinel-1 SAR (synthetic aperture radar) measurements, natively at a 10m spatial resolution; 
(2) Sentinel-2 multispectral measurements (RGB, infrared, etc.), natively at a 10m spatial resolution;
(3) MODIS land cover labels: a classification of each pixel into one of several land cover categories (water bodies, forest, urban land, etc.), natively at a 500m spatial resolution, but upsampled to a 10m resolution.
%
% \end{enumerate}
% \begin{enumerate}
%     \item Sentinel-1 SAR (synthetic aperture radar) measurements, natively at a 10m spatial resolution
%     \item Sentinel-2 multispectral measurements (RGB, infrared, etc.), natively at a 10m spatial resolution
%     \item 
%     MODIS land cover labels: a classification of each pixel into one of several land cover categories (water bodies, forest, urban land, etc.), natively at a 500m spatial resolution, but upsampled to a 10m resolution.
% \end{enumerate}
%
The MODIS labels contain 4 layers, each corresponding to a unique land-cover categorization scheme. Note that these MODIS labels are gathered at a 500m resolution, so while the images themselves are 256x256 pixels, each label ``block'' occupies quite a large region of each 256x256 image (see Figure ~\ref{fig:sample-images}). This makes model fitting nontrivial, because the labels themselves are coarse, noisy, and may not exactly correspond with real-life ground truth.
According to \cite{2019RSEnv.222..183S}, the overall accuracies of the MODIS layers are about 67\% (IGBP), 74\% (LCCS land cover), 81\% (LCCS land-use), and 87\% (LCCS surface hydrology), respectively. This should be kept in mind when using the land cover data to measure model performance, since these figures will constitute an upper bound for the model's out-of-sample accuracy.

%\vspace{-0.4in}
\section{Model Generalization}

This report focuses on the cross-continent and cross-season land-cover classification tasks: given a set of imagery from one continent or season, we want to train a pixel-level land-cover classification model on that imagery such that it can accurately predict land-cover on each pixel in a different season/continent. In other words, we formulate the problem as a semantic segmentation task, where the training and testing set represent different seasons/continents.

For our labels, we use a slightly consolidated version of the LCCS land-use (LU) classification scheme as described in~\cite[Section~5]{schmitt2019sen12ms}. Whereas the original LCCS LU scheme contains 11 unique classes, the consolidated version combines ``Open Forests'' and ``Forest/Cropland Mosaics'' into a single ``Open Forests'' class, as well as ``Natural Herbaceous'', ``Herbaceous Croplands'', and ``Natural Herbaceous/Cropland Mosaics'' into a single ``Natural Herbaceous'' class, resulting in a total of 8 consolidated classes.

\subsection{Model Architecture and Experimental Setup}

We selected a Fully-Convolutional DenseNet model, a type of convolutional neural network, for this semantic segmentation task based on its success in previous work~\cite{jgou2016layers}. The input features and target classes are shown in Table ~\ref{tbl:model-input-output-description}, and the model configuration is shown in Table ~\ref{tbl:model-config}. (See \cite{schmitt2019sen12ms} for a more detailed definition of each of input feature.) For the forward pass, we pass in the entire 256x256 patch as input, and produce a 256x256x8 land-use mask as output. The last (channel) dimension represents $P(y|x)$, a pixel-level probability estimate over the 8 possible output classes, as computed by a softmax transformation of the output logits.

\begin{table}
 \centering
 \begin{minipage}{0.55\textwidth}
     \centering 
     \begin{tabular}{|p{0.45\textwidth}|p{0.45\textwidth}|} 
     \hline
     \multicolumn{1}{|c|}{Sentinel-2 Input Bands (10)} & \multicolumn{1}{|c|}{Consolidated LU Classes (8)} \\
     \hline
     \begin{itemize}[leftmargin=4mm]
        \item RGB: \texttt{blue}, \texttt{green}, \texttt{red}
        \item Red Edge: \texttt{re1}, \texttt{re2}, \texttt{re3}
        \item Near Infrared: \texttt{nir1}, \texttt{nir2}
        \item Short-Wave Infrared:\newline \texttt{swir1}, \texttt{swir2}
    \end{itemize} & \begin{itemize}[leftmargin=4mm]
        \item Dense Forests
        \item Open Forests
        \item Herbaceous
        \item Shrublands
        \item Urban and Built-Up Lands
        \item Permanent Snow \& Ice
        \item Barren
        \item Water Bodies 
    \end{itemize} \\
    \hline
    \end{tabular}
    \caption{Description of the model inputs and outputs.}
    \label{tbl:model-input-output-description}
  \end{minipage}\hfill
  \begin{minipage}{0.4\textwidth}
     \centering
     \begin{tabular}{|p{0.9\textwidth}|} 
     \hline
     \multicolumn{1}{|c|}{Fully-Convolutional DenseNet Configuration} \\
     \hline
     \begin{itemize}[leftmargin=4mm]
        \item Input shape: $256 \times 256 \times 10$
        \item Output shape: $256 \times 256 \times 8$
        \item Batch size: 4
        \item Loss: Categorical cross-entropy
        \item Initial learning rate: 0.0001
        \item LR schedule: ReduceOnPlateau
        \item Optimizer: Adam
        \item Num. dense blocks: 3
    \end{itemize} \\
     \hline
    \end{tabular}
    \caption{FC-DenseNet model configuration.}
    \label{tbl:model-config}
  \end{minipage}
  \vspace{-10mm}
\end{table}

%\subsection{Experimental Setup}

To measure cross-continent model generalization, we train a model on all the scenes from within a given continent (withholding 20\% of the within-continent scenes as a validation set for early stopping), and then evaluate that model on each scene from all other continents. We then compare the model's performance on the withheld within-continent scenes to the model's performance on out-of-continent scenes.
The cross-season case is analogous to the cross-continent case: instead of training on all the scenes from a given continent, we instead train on all the scenes from a given season, and then evaluate the performance of each model on scenes from all other seasons.

The code for these experiments can be found at \url{https://github.com/lucashu1/land-cover}.

\subsection{Results}

\begin{figure}
    \centering
    \includegraphics[width=0.85\textwidth]{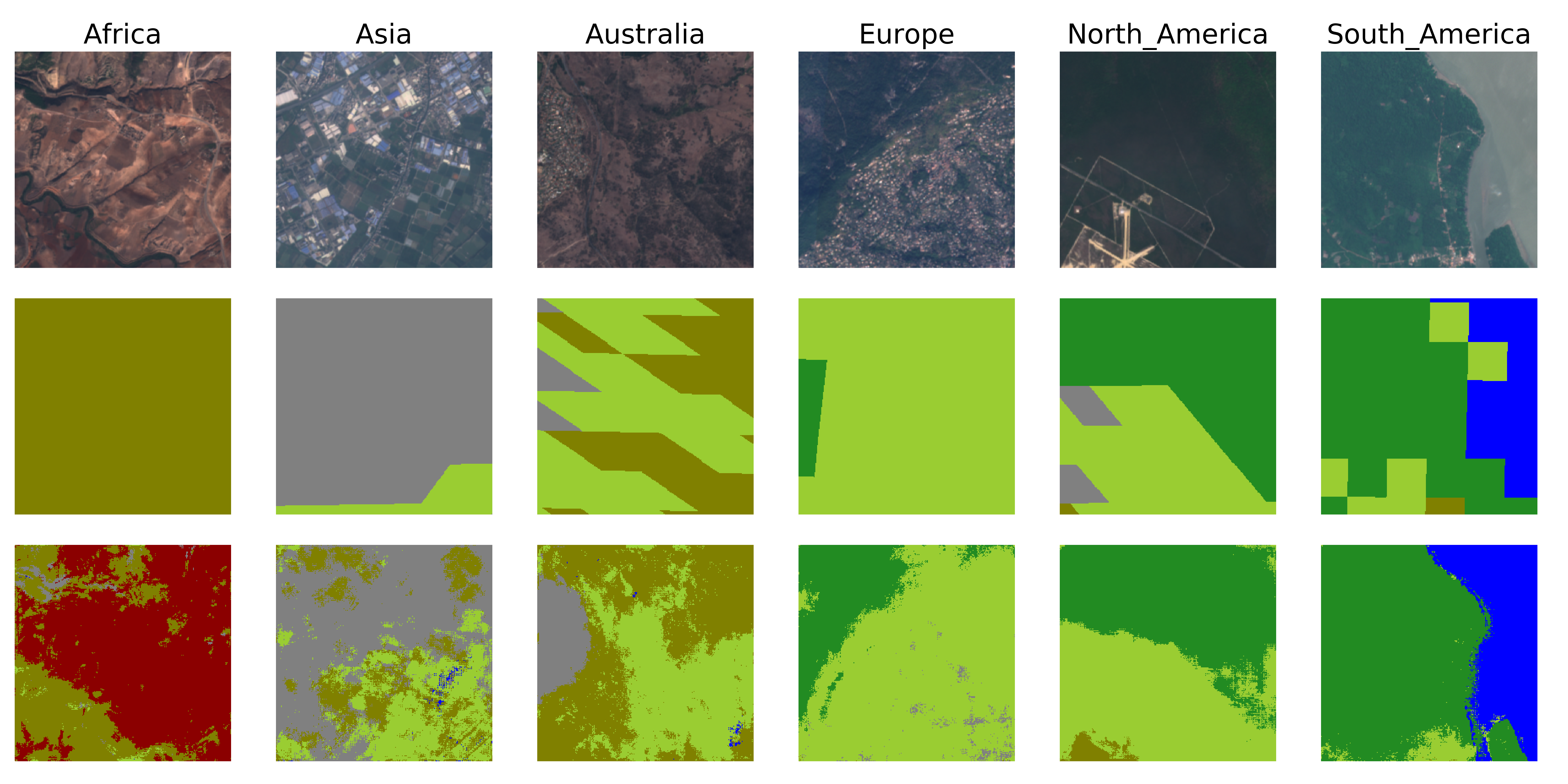}
    \begin{minipage}{0.85\textwidth}
    \centering
    \begin{tabular}{llll}
    \fcolorbox{black}[HTML]{8B0000}{\rule{0pt}{3pt}\rule{3pt}{0pt}}\quad~{Barren} &
    \fcolorbox{black}{white}{\rule{0pt}{3pt}\rule{3pt}{0pt}}\quad~{Permanent Snow \& Ice} &
    \fcolorbox{black}[HTML]{0000FF}{\rule{0pt}{3pt}\rule{3pt}{0pt}}\quad~Water &
    \fcolorbox{black}[HTML]{808080}{\rule{0pt}{3pt}\rule{3pt}{0pt}}\quad~{Urban and Built-Up Lands} \\
    \fcolorbox{black}[HTML]{228B22}{\rule{0pt}{3pt}\rule{3pt}{0pt}}\quad~{Dense Forests} &
    \fcolorbox{black}[HTML]{9ACD32}{\rule{0pt}{3pt}\rule{3pt}{0pt}}\quad~{Open Forests} &
    \fcolorbox{black}[HTML]{DAA520}{\rule{0pt}{3pt}\rule{3pt}{0pt}}\quad~{Natural Herbaceous} &
    \fcolorbox{black}[HTML]{808000}{\rule{0pt}{3pt}\rule{3pt}{0pt}}\quad~Shrublands \\
    \end{tabular}
    \end{minipage}
    \captionsetup{justification=centering}
    \caption{Sample in-continent land cover predictions. Top: Sentinel-2 RGB. Middle: Consolidated LCCS Land-Use labels. Bottom: Predicted land-use.}
    \label{fig:sample-preds}
    \vspace{-5mm}
\end{figure}

Accuracy results for the FC-DenseNet experiments are shown in Tables \ref{tbl:cross-continent-results} and \ref{tbl:cross-season-results}. Each row shows the results for a single model trained on a given continent/season, as well as the number of scenes on which the model was trained (excluding scenes used for validation/early stopping). Each column shows the results for each model when evaluated on a given continent/season.
Each value in the table represents an overall accuracy metric (i.e. the percentage of pixels in each image that were correctly classified) averaged over each scene in the per-continent/per-season evaluation set. Standard deviations for this metric, also computed over each scene, are shown as well. The diagonal entries represent the model's accuracy on the 20\% validation set from within the same continent/season. Sample in-continent predictions are shown in Figure ~\ref{fig:sample-preds}.

\begin{table}
\centering
 \begin{tabular}{|p{2cm}|p{1.7cm}|p{1.7cm}|p{1.7cm}|p{1.7cm}|p{1.7cm}|p{1.7cm}|} 
 \hline
 \backslashbox[2.3cm]{Train}{Test} & Africa & Asia & Australia & Europe & N. America & S. America \\
 \hline
 Africa & $\mathbf{0.643 \pm 0.132}$ & $0.460 \pm 0.228$ & $0.396 \pm 0.281$ & $0.423 \pm 0.167$ & $0.471 \pm 0.184$ & $0.589 \pm 0.163$ \\
 \hline
 Asia & $0.472 \pm 0.274$ & $\mathbf{0.694 \pm 0.179}$ & $0.501 \pm 0.273$ & $0.576 \pm 0.226$ & $0.495 \pm 0.220$ & $0.536 \pm 0.234$ \\
 \hline
 Australia & $0.425 \pm 0.290$ & $0.380 \pm 0.237$ & $\mathbf{0.682 \pm 0.104}$ & $0.419 \pm 0.207$ & $0.504 \pm 0.199$ & $0.518 \pm 0.192$ \\
 \hline
 Europe & $0.510 \pm 0.285$ & $0.630 \pm 0.256$ & $0.452 \pm 0.324$ & $\mathbf{0.792 \pm 0.070}$ & $\mathbf{0.612 \pm 0.212}$ & $0.588 \pm 0.208$ \\
 \hline
 N. America & $0.548 \pm 0.302$ & $0.630 \pm 0.230$ & $0.477 \pm 0.348$ & $0.674 \pm 0.169$ & $0.529 \pm 0.283$ & $0.658 \pm 0.173$ \\
 \hline
 S. America & $0.516 \pm 0.325$ & $0.502 \pm 0.234$ & $0.446 \pm 0.341$ & $0.578 \pm 0.232$ & $0.583 \pm 0.215$ & $\mathbf{0.757 \pm 0.111}$ \\
 \hline
\end{tabular}
\caption{Cross-continent FC-DenseNet accuracy results.}
\label{tbl:cross-continent-results}
\vspace{-8mm}
\end{table}

\begin{table}
\centering
\begin{tabular}{|l|*{4}{c|}} 
 \hline
 \backslashbox{Train}{Test} & Spring & Summer & Fall & Winter \\
 \hline
 Spring & $0.518\pm 0.230$ & $0.537\pm 0.200$ & $0.569\pm 0.205$ & $0.570\pm 0.219$ \\
 \hline
 Summer & $\mathbf{0.532 \pm 0.285}$ & $\mathbf{0.618 \pm 0.110}$ & $\mathbf{0.582 \pm 0.250}$ & $0.546 \pm 0.255$ \\
 \hline
 Fall & $0.508 \pm 0.280$ & $0.484 \pm 0.223$ & $0.495 \pm 0.332$ & $\mathbf{0.595 \pm 0.216}$ \\
 \hline
 Winter & $0.494 \pm 0.284$ & $0.521 \pm 0.225$ & $0.538 \pm 0.258$ & $0.522 \pm 0.258$ \\
 \hline
\end{tabular}
\caption{Cross-season FC-DenseNet accuracy results.}
\label{tbl:cross-season-results}
\vspace{-10mm}
\end{table}

%\subsection{Discussion}

When attempting to explain these results, one should remember that the MODIS land-use labels themselves are quite noisy (only about 81\% accurate, as mentioned in Section~\ref{dataset}), which may contribute significantly to the high variance in the observed results. Furthermore, since the MODIS labels only operate on a 500m resolution, 50 times coarser than the 10m resolution at which we evaluated each of the land cover models, it should not be surprising to see a significant amount of noise in the experimental results as well.
Looking at the results on the whole, though, some general trends do start to emerge. In both the cross-continent and cross-season cases, it appears that overall accuracy is generally highest when evaluated on the same continent/season on which the model itself was trained. In other words, \textit{out-of-continent/out-of-season} accuracy tends to be lower than \textit{in-continent/in-season} accuracy.

More generally, there appears to be a high degree of variability in model performance across different continents/seasons. This suggests that just because a model accurately predicts land-use classes in one continent or season does not mean that the model will accurately predict land-use classes in a \textit{different} continent or season.
These differences in model performance can sometimes be quite drastic. The DenseNet model that was trained on scenes from Europe, for example, had a mean accuracy of 0.705 when evaluated on the validation scenes from within Europe, but only had a mean accuracy of 0.424 when evaluated on scenes from within Africa. The variations in cross-season model performance seem to be somewhat smaller than the variations in cross-continent model performance, but are still significant enough to be non-negligible.
One possible explanation for the large differences in cross-continent performance --- in addition to the inherent noisiness in the MODIS labels --- may be that different land-use classes may in fact look quite different in different continents. When training a model on scenes from only one continent, the model may learn to overfit to the particular features of that one continent, making the model less likely to generalize well to new regions. If this is indeed the case, then it seems useful to explore the efficacy of various regularization techniques to help prevent this sort of \textit{regional/seasonal overfitting}.

\section{Landscape Visualization}

Next, we investigate the underlying reasons behind this lack of geospatial generalization in deep learning-based land cover classification models. We focus on the problem of cross-continent generalization, due to the discrepancies in landscapes between different continents. Specifically, we propose a clustering-based technique to visualize the differences in landscapes between different regions.

\subsection{Per-Continent Input Distributions}

\begin{figure}[!htb]
    \centering
    \includegraphics[width=0.9\textwidth]{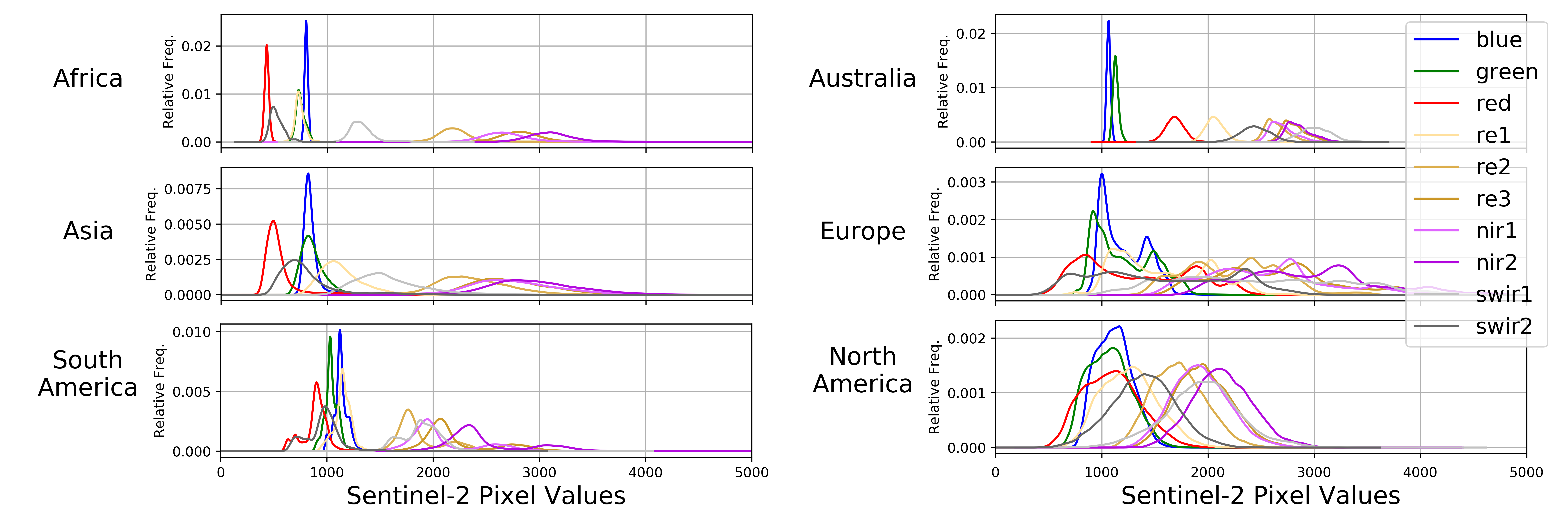}
    \vspace{-4mm}
    \caption{Sentinel-2 band distributions.}
    \label{fig:band_dist}
    % \vspace{-3mm}
\end{figure}

We hypothesize that a key factor that makes geospatial generalization difficult is that each continent may have significantly different input distributions, indicating the presence of different landscapes. For example, South America may contain more satellite imagery of rainforest landscapes than Europe does, making it difficult for a model trained on images from South America (and therefore trained on more images of rainforest landscapes) to perform well when classifying land cover in Europe. To test this, we first investigate the differences in images between continents. In Figure ~\ref{fig:band_dist}, we show Kernel Density Estimation plots for all Sentinel-2 bands across different continents.
Comparing the KDE plots between continents, it appears that each continent seems to display a relatively unique ``signature'' as characterized by the distribution over each input band. Australia, for example, has a sharp spike in the ``red'' band around 1600, whereas Europe has a much more uniform distribution in its ``red'' band values.
These differences suggest that by looking at the band distribution of any given image and comparing it to the band values observed in each continent, we may be able to infer which continent -- or which type of landscape -- that image is more likely to belong to. In other words, looking at the band distribution of a given image may help us determine if a given image is ``in-distribution'' or not, relative to other images in each continent and/or landscape.

\subsection{Landscape Clustering}

\begin{figure}[!htb]
    \centering
    \includegraphics[width=0.8\textwidth]{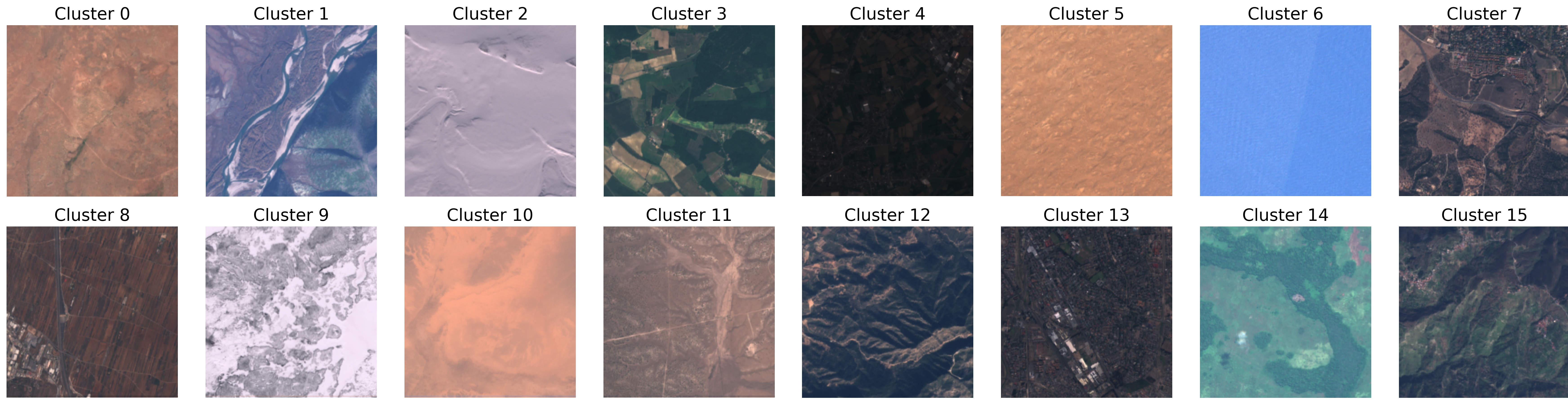}
    \vspace{-3mm}
    \caption{Representative images from each cluster.}
    \label{fig:cluster_images}
    \vspace{-3mm}
\end{figure}

To help further visualize the idea that the same classes look different depending on the continent, we cluster all images in the entire SEN12MS dataset according to the means of each Sentinel band over an entire patch.
Since there are 10 input bands in total, we represent each image by a 10-dimensional vector. We then fit a K-Means model (with $K = 16$ clusters) on these 10-dimensional vectors. The most representative (i.e. closest to centroid) images for each cluster are shown in Figure ~\ref{fig:cluster_images}.

% \begin{figure}[!htb]
%     \centering
%     \includegraphics[width=0.6\textwidth]{figures/image-cluster-model-2.png}
%     \caption{Image cluster k-means model.}
%     \label{fig:cluster_model}
% \end{figure}

\begin{figure}[!htb]
    \centering
    \begin{minipage}{0.575\textwidth}
        \centering
        \includegraphics[width=0.9\textwidth]{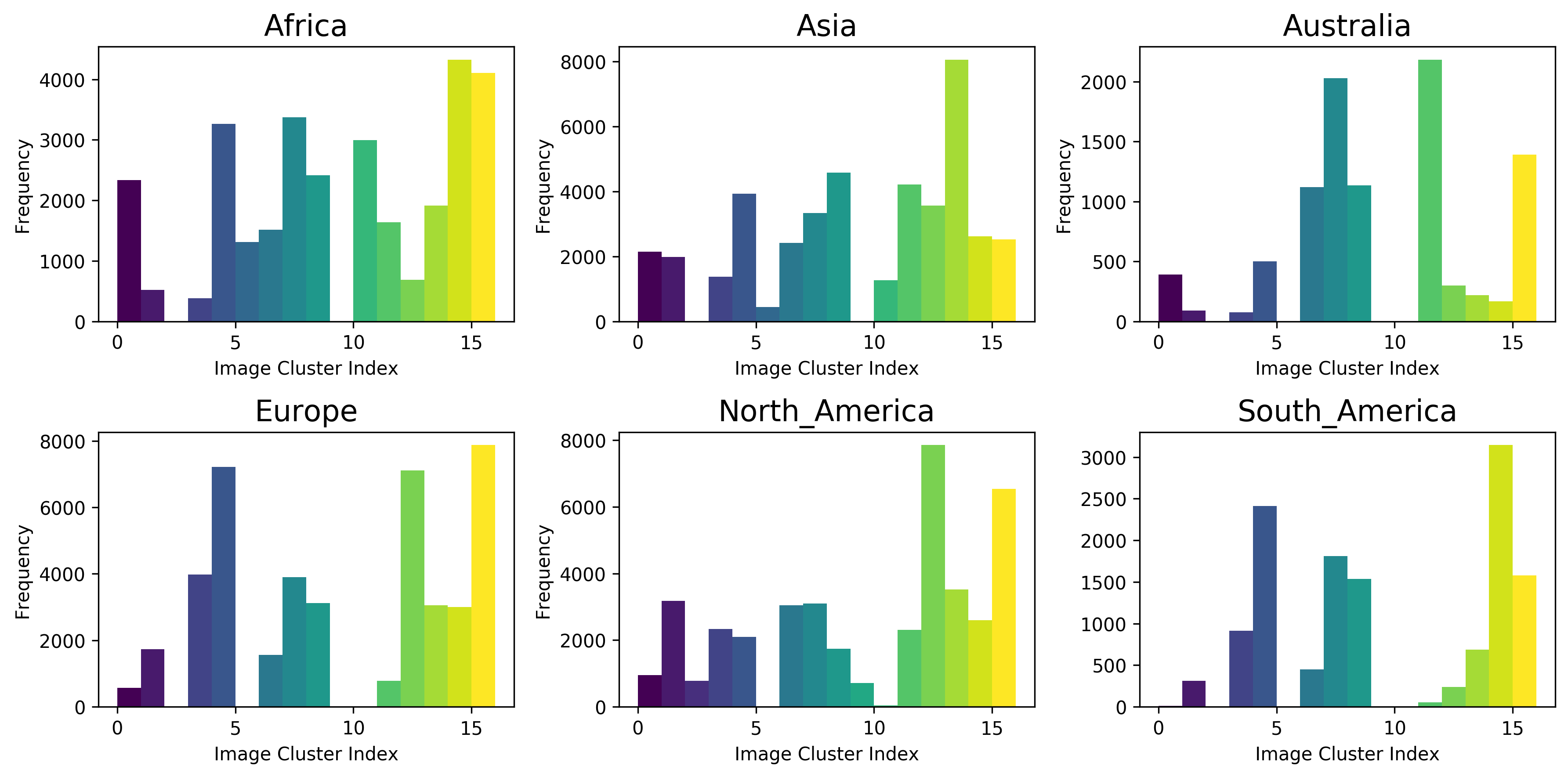}
        \caption{Image cluster histograms, per-continent.}
        \label{fig:per_continent}
    \end{minipage}\hfill
    \begin{minipage}{0.4\textwidth}
        \centering
        \includegraphics[width=0.9\textwidth]{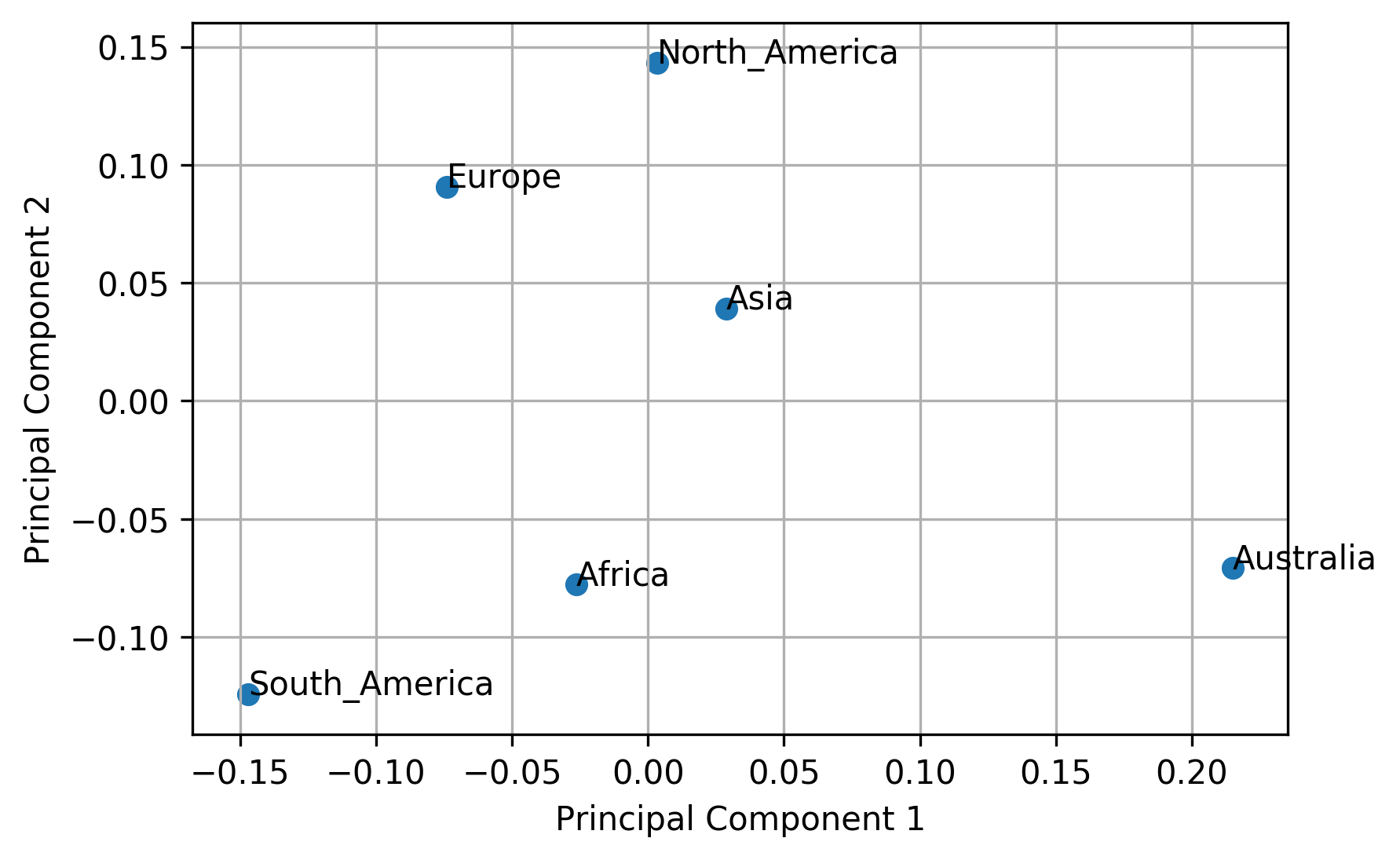}
        \caption{PCA visualization of cluster histograms.}
        \label{fig:pca}
    \end{minipage}
    \vspace{-3mm}
\end{figure}

As one can see, each cluster seems to represent a unique type of landscape. For example, Clusters 0, 5, and 10 represent more desert-looking landscapes, whereas Clusters 2 and 9 represent snow and/or tundra. If different continents do have different types of landscapes, then we should expect that the distribution over these clusters will differ between continents. After finding the histogram over these image clusters for each continent, we can show that this is indeed the case (see Figure ~\ref{fig:per_continent}). South America, for example, seems to have a higher prevalence of Cluster 14, which appears to be a rainforest-type landscape. Australia seems to have a higher prevalence of Cluster 11, which seems to represent a more arid, outback-type landscape.

After normalizing each continent's cluster-histograms, we can also apply PCA on these histograms to visualize which continents are most similar or different to each other (see Figure ~\ref{fig:pca}). Looking at this PCA plot, we can see that Africa and South America tend to have similar landscapes; Europe and North America, too, are relatively similar. This may explain why, in Section 3.3, the model trained on Europe was also able to perform well when evaluated on North America, and why the model trained on South America was also able to perform well in Africa. Australia's landscapes are unique relative to all other continents, which may also explain why no other continent-models were able to perform well when applied to Australia. Lastly, Asia appears to occupy a middle ground between all continents.

\begin{figure}[!htb]
    \centering
    \includegraphics[width=0.9\textwidth]{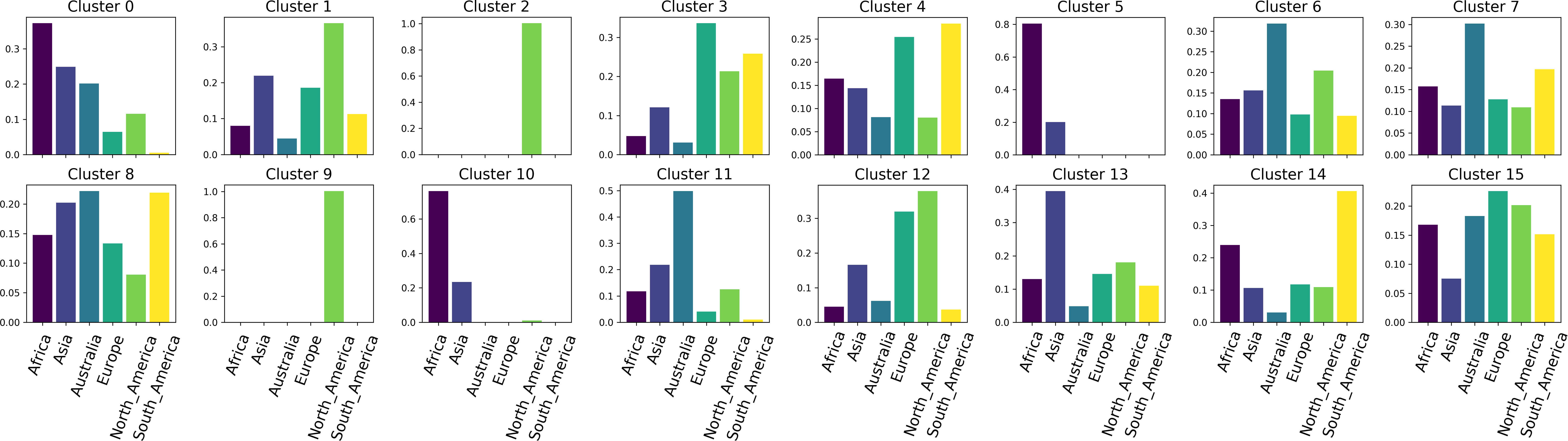}
    \vspace{-2mm}
    \caption{$P (Continent | Cluster)$, normalized.}
    \label{fig:p_continent_cluster}
    \vspace{-8mm}
\end{figure}

\sloppy Finally, we can use these per-continent histograms to infer ${P(continent | cluster)}$. Since the dataset contains a different number of images per continent, we normalize these probabilities as follows: ${P(continent | cluster) \propto P(cluster | continent) / P(continent)}$. The resulting probabilities are shown in Figure ~\ref{fig:p_continent_cluster}.
As one can see, certain landscape clusters are indeed far more likely to belong to certain continents than to others. For example, Clusters 5 and 10 (desert landscapes) almost exclusively belong to either Africa or Asia. Cluster 9 and 2 (snow landscapes) exclusively belong to North America.
Since certain landscape clusters can be indicative of belonging to a certain continent, we may be able to take advantage of these landscape distributions to produce models that perform better in certain continents -- or at least, to know if a model trained on a certain set of landscape images is capable of performing well on a given continent.

\section{Conclusion}

In order to evaluate how well geospatial models trained on one season/continent generalize to other seasons/continents, we trained land-use classification models on data from specific seasons/continents, and evaluated these models on data from different seasons/continents. Experimental results show that out-of-continent/out-of-season accuracy tends to be lower than in-continent/in-season accuracy, and that there is often a high amount of variability in model performance across different continents/seasons.
We then apply clustering methods to patches of satellite imagery based on mean Sentinel band values, demonstrating that different continents do comprise visually different landscapes, which may explain the difficulties in geospatial generalization across continents.

This leads to three main insights regarding the generalizability of geospatial models:

\begin{enumerate}
    \item Practitioners should exercise caution when naively applying geospatial models trained on one season or region to new seasons or regions.
    \item Training geospatial models on data from a diverse set of seasons and/or continents may be necessary to promote model generalizability.
    \item When using an existing model to perform land-cover classification on a new region, it is often helpful to know whether the model was trained on imagery from a similar landscape -- e.g. by organizing images into clusters, and seeing if the new images match  with the cluster indices of the images from the training set. If not, it may be necessary to fall back to a simpler model, rather than naively applying the existing model to the new region.
    
\end{enumerate}

Future work is necessary to determine what techniques (regularization, data augmentation, unsupervised pre-training, etc.) may be required to promote cross-continent and cross-season generalization, and to reduce performance degradation when predicting on out-of-sample remote sensing data in general.

%%
%% The acknowledgments section is defined using the "acks" environment
%% (and NOT an unnumbered section). This ensures the proper
%% identification of the section in the article metadata, and the
%% consistent spelling of the heading.
\begin{acks}
The authors would like to thank Michael Schmitt and Chunping Qiu for their help in providing the starter code for both the FC-DenseNet land-use classification models~\cite{schmitt2019sen12ms}.
\end{acks}

%%
%% The next two lines define the bibliography style to be used, and
%% the bibliography file.
\bibliographystyle{ACM-Reference-Format}
\bibliography{citations}

%%
%% If your work has an appendix, this is the place to put it.

\end{document}